# DESIGN AND DEVELOPMENT OF ARTIFICIAL NEURAL NETWORKING (ANN) SYSTEM USING SIGMOID ACTIVATION FUNCTION TO PREDICT ANNUAL RICE PRODUCTION IN TAMILNADU

S.Arun Balaji[1] and K.Baskaran[2]

[1]PhD scholar, Computer Science & Engineering, Karpagam University, Coimbatore, India.
Email: arunbalaji1983@yahoo.co.in

[2]PhD Research Director and Associate Professor, Dept. of Computer Science & Engineering and Information Technology, Government College of Technology (GCT), Coimbatore, India.
Email: baski_101@yahoo.com

## ABSTRACT

*Prediction of annual rice production in all the 31 districts of Tamilnadu is an important decision for the Government of Tamilnadu. Rice production is a complex process and non linear problem involving soil, crop, weather, pest, disease, capital, labour and management parameters. ANN software was designed and developed with Feed Forward Back Propagation (FFBP) network to predict rice production. The input layer has six independent variables like area of cultivation and rice production in three seasons like Kuruvai, Samba and Kodai. The popular sigmoid activation function was adopted to convert input data into sigmoid values. The hidden layer computes the summation of six sigmoid values with six sets of weightages. The final output was converted into sigmoid values using a sigmoid transfer function. ANN outputs are the predicted results. The error between original data and ANN output values were computed. A threshold value of $10^{-9}$ was used to test whether the error is greater than the threshold level. If the error is greater than threshold then updating of weights was done all summations were done by back propagation. This process was repeated until error equal to zero. The predicted results were printed and it was found to be exactly matching with the expected values. It shows that the ANN prediction was 100% accurate.*

## 1. INTRODUCTION

Rice is the stable food for Tamil Nadu. Prediction of annual rice production in all the 31 districts of Tamilnadu is an important decision for the Government of Tamilnadu so as to plan for importing rice from other state or exporting rice to other states to meet the food security. One of the pillars of success of the Government depends on planning the availability of rice. Rice production is based on the soil type, rainfall, atmospheric temperature, sunshine intensity, duration of sunshine, manure applied, inorganic fertilizers applied, weed control, application of timely and sufficient irrigation water, outbreak of pests and diseases etc. Hence, rice production is a complex process and hence, it is difficult to predict with the available data using serial computations and serial algorithms. Some scientists attempted the prediction systems using different modeling and simulations packages as a serial processing approach and meet varied levels of success. To the best of the knowledge of the authors, nobody has made an attempt in predicting rice production in Tamilnadu by building an ANN technique. Literature review states that ANN is used to solve more complex problems (non linear problems) based on parallel processing approach rather than the serial processing.

An Artificial Neural Network (ANN) is an information processing paradigm that is inspired by the way biological nervous systems, such as the brain process information. The key element of this paradigm is the novel structure of the information processing system. It is composed of a large number of highly interconnected processing elements (neurons) working in unison to solve specific problems. ANNs, like people, learn by example. An ANN is configured for a specific application, such as pattern recognition or data classification, through a learning process. Learning in biological systems involves adjustments to the synaptic connections that exist between the neurons. This is true of ANNs as well. Neural networks are clusters of neurons that are interconnected to process information.

## 1.1 Benefits of ANN

Already ANN was used to predict the weather forecasting problems and stock market behaviors. Hence, the present paper is unique in the sense ANN is used to predict rice production in Tamilnadu. At the same time, this research is needed for the food planning for Tamilnadu state to solve a complex process using parallel processing approach. Several benefits of using ANN for predicting rice production in Tamilnadu are shown below:

- ANN is a powerful modeling technique capable of solving non linear and complex process of rice production using parallel processing approach
- ANN learns itself from the structure of training dataset and predicts the future data.
- ANN uses dimensionless numbers in its input and output variables.
- User feed input and get output without thinking about the dimensionality in computation inside the hidden layer.

The overall objective of the present research is to "Design and Development of Artificial Neural Networking (ANN) system using sigmoid activation function to predict annual rice production in Tamilnadu".

## 2. REVIEW OF LITERATURE

### 2.1 Biology of Human Intelligence

The brain and nervous system are enormously complex. The brain itself is composed of billions of nerve cells. The orchestration of all of these cells to allow people to sing, dance, write, talk, and think. Neuroscientist [1] calls the brain the *great integrator*. The brain does a wonderful job of pulling information together. The brain integrates all functions of the world including sounds, sights, touch, taste, genes and environment. Neurons are the nerve cells that actually handle the information processing function. The human brain contains about 100 billion neurons. The average neuron is as complex as a small computer and has as many as 10,000 physical connections with other cells.

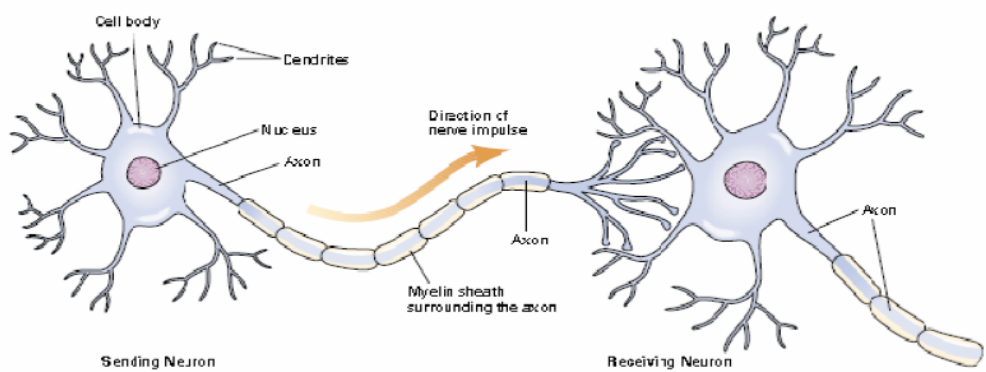

Figure 1.  Connection between one neuron to other neuron.
(Source: John W. Santrock, University of Texas, Dallas, 2005)

To have even the merest thought requires millions of neurons acting simultaneously [1]. Nerve cells, chemicals, and electrical impulses work together to transmit information at speeds of up to 330 miles per hour. As a result, information can travel from our brain to our hands (or vice versa) in a matter of milliseconds.

Neurons are specialized to handle different information processing functions. All neurons do have some common characteristics. Most neurons are created very early in life, but their shape, size, and connections can change throughout the life span. They are not fixed and immutable but can change. Every neuron has a cell body, dendrites, and axon (see figure 1).

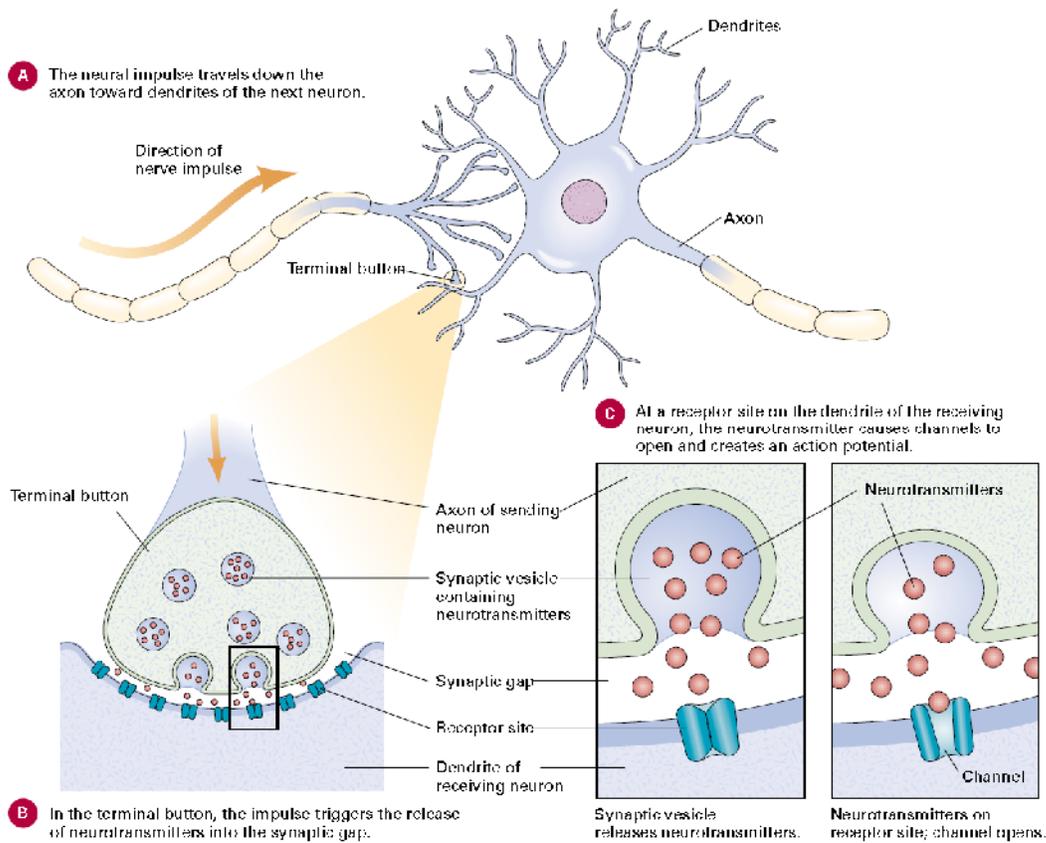

Figure 2: The neural impulse travels down the axon toward dendrites of the next neuron.
(Source: John W. Santrock, University of Texas, Dallas, 2005)

The **cell body** contains the nucleus, which directs the manufacture of substances that the neuron needs for growth and maintenance.

**Dendrites** receive and orient information toward the cell body. One of the most distinctive features of neurons is the tree-like branching of their dendrites. Most nerve cells have numerous dendrites, which increase their surface area, allowing each neuron to receive input from many other neurons.

The **axon** is the part of the neuron that carries information away from the cell body toward other cells. Although very thin (1/10,000th of an inch), axons can be very long, with many branches. In fact, some extend more than three feet—all the way from the top of the brain to the base of the spinal cord.

Covering all surfaces of neurons, including the dendrites and axons, are very thin cellular membranes that are much like the surface of a bubble. The neuronal membranes are semipermeable, meaning that they contain tiny holes or *channels* that allow only certain substances to pass into and out of the neurons.

A **myelin sheath,** a layer of fat cells, encases and insulates most axons. By insulating axons, myelin sheaths speed up transmission of nerve impulses. Figure 2 shows how neural impulse travel from sending neuron to receiving neuron.

### 2.1.1 The Neural Impulse

A neuron sends information through its axon in the form of brief impulses, or waves, of electricity. To transmit information to other neurons, a neuron sends impulses ("clicks") through its axon to the next neuron. By changing the rate and timing of the signals or "clicks," the neuron can vary its message.

### 2.1.2 Working of Synapses and Neurotransmitters:

a) The axon of the sending neuron meets dendrites of the receiving neuron.
b) This is an enlargement of one synapse, showing the synaptic gap between the two neurons, the terminal button, and the synaptic vesicles containing a neurotransmitter.
c) There is an enlargement of the receptor site.

It is to be noted that how the neurotransmitter opens the channel on the receptor site, triggering the neuron to fire.

## 2.2 ANN architecture

It was stated by [2] that a neural network is a massively parallel distributed processor that has a natural propensity for storing experiential knowledge and making it available for use. Artificial neural networks are computers whose architecture is modeled after the brain. It resembles the brain in two respects: knowledge is acquired by the network through a learning process and inter-neuron connection strengths known as synaptic weights are used to store the knowledge. They typically consist of many hundreds of simple processing units, which are wired, together in a complex communication network. Each unit or node is a simplified model of a real neuron, which fires (sends off a new signal) if it receives a sufficiently strong input signal from the other nodes to which it is connected. The strength of these connections may be varied to enable the network to perform different tasks corresponding to different patterns of node firing activity. The basic element of a neural network is the perceptron as shown in figure 3 below.

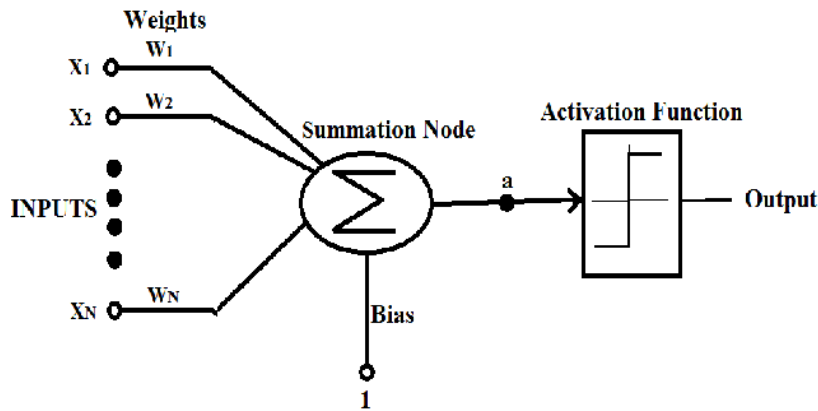

Figure 3: A simple perceptron Neural Network

## 2.2.1 Sigmoid Function

A sigmoid function is a mathematical function having an "S" shape (sigmoid curve). Sigmoid function is given by the following formula:

$$S(t) = \frac{1}{1+e^{-t}}.$$

Sigmoid functions are very similar to the input-output relationships of biological neurons, although not exactly the same. Sigmoid function exhibits smoothness and has the desired asymptotic properties. The sigmoid curve is shown in figure 4.
As t goes to minus infinity, S(t) goes to 0.
As t goes to infinity, S(t) goes to 1.
As t =0, S(t) =0.5

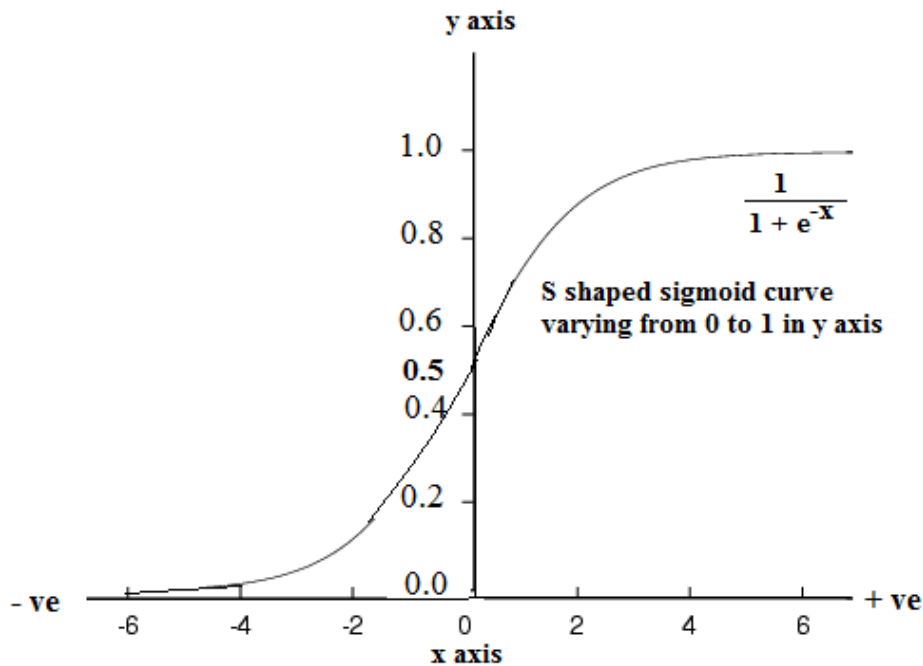

Figure 4: Sigmoid activation function varies from 0 to 1 as a S shaped curve

## 2.2.2 Error Correction Learning

Error-Correction Learning is used with supervised learning. It is the technique of comparing the system's ANN output to the desired output value. The error is used to train the system for better performance. The error values can be used to directly adjust thee weights, using an algorithm such as the back propagation algorithm. If the system output is ANN output and the desired system output is known actual data supplied. As per [3], the error can be computed using the formula given below:

$$\text{error} = \frac{\text{actual output} - \text{ANN output}}{\text{actual output}}$$

Compare the error with threshold value of $10^{-9}$. If the error was greater than threshold value then calculate the updated weights and compute the summation using back propagation. This process is repeated until error is zero.

## 2.3 ANN based applications

According to [4], it was attempted to forecast groundwater level of a watershed using ANN and Fuzzy Logic. A three-layer feed-forward ANN was developed using the sigmoid function and the back propagation algorithm. It was observed that ANNs perform significantly better than Fuzzy Logic.

According to [5], prediction of rainfall in Chennai using back propagation neural network model was carried out, the mean monthly rainfall is predicted using ANN model. The model can perform well both in training and independent periods.

According to [6] there was a comparison of Dynamic Vs Static neural network models for rainfall forecasting. AI based forecasting architectures using Multi-Layer Perceptron Neural Networks (MLPNN) and Adaptive Neuron-Fuzzy Inference Systems (ANFIS could be suitable for modeling the temporal dimension of the rainfall pattern and provided better forecasting accuracy.

On the basis of humidity, dew point and pressure in India [7] has used the back propagation neural network model for predicting the rainfall. In the training it was obtained that the prediction accuracy was 99.79% in training and 94.28% in testing. From these results, it was concluded that rainfall can be predicted in future using the back propagation neural network.

Rainfall-runoff prediction based on Artificial Neural Network was studied by [8] and found that Artificial Neural Network method is more appropriate and efficient to predict the river runoff than classical regression model.

The ANN model for Long-Range Meteorological Parameters Pattern Recognition over the Smaller Scale Geographical Region was developed by [9] and the performances of this model in pattern recognition and prediction have been found to be extremely good.

ANN was used in a new experiment [10] of short term temperature forecasting (STTF) and found that multi layer perceptron network has the minimum forecasting error and can be considered as a good method to model the STTF systems

According to [11] work was carried out to find out best hidden layer size for three layered neural net work in predicting monsoon rainfall in India, and it was found that eleven-hidden-nodes in three-layered neural network had more efficacy than asymptotic regression in the present forecasting.

ANN has the working features like human brain [12] and it has many desirable characteristics. ANN has the following features:

a.   massive parallelism
b.   distributed representation and computation
c.   learning ability
d.   generalization ability
e.   Adaptivity to complex problems, without knowing dimensions of data
f.   inherent contextual information processing
g.   fault tolerance
h.   low energy consumption

NNs are universal approximators that can map any non linear functions [13]. NNs are powerful tools for solving classifications and forecasting. NNs are less sensitive to error term assumptions and can tolerate noise and chaotic components better than other methods.

Bendiktsoon et al. [14] compared neural networks and statistical approaches, together, with a multispectral data classification. They noted that conventional multivariate classification methods cannot be used in processing multisource spatial data. This is due to different distribution properties and measurement scales.

Heermann and Khazenie [15] compared neural networks with more classical statistical methods. Heerman and Khazenie's study emphasized the analysis of larger data sets with back propagation methods, in which error is distributed throughout the network. They concluded that the back propagation network could be easily modified to accommodate more features or to include spatial and temporal information.

Hepner et al. [16] compared the use of neural network back propagation with a supervised maximum likelihood classification method, using a minimum training set. The results showed that a single training site per class of neural network classification was comparable to a four training site per class of conventional classification. The result demonstrated that the neural network method offered a potentially more robust approach to land cover classification than conventional image classification methods.

Amini, J. [17] used, a Multi Layer Perceptron (MLP) network with a Back Propagation (BP) algorithm is used for the classification of IRS-1D satellite images. A network with an optimum learning rate is proposed. The MLP consists of neurons that are arranged in multiple layers with connections only between nodes in the adjacent layers by weights. The layer where the input information is presented is known as the input layer and the layer where the processed information is retrieved is called the output layer. All layers between the input and output layers are known as hidden layers. For all neurons in the network, except the input layer neurons, the total input of each neuron is the sum of the weighted outputs of the neurons in the previous layer. Each neuron is activated with input to the neuron and by the activation function of the neuron [18].

## 3. METHODOLOGY

The methodologies adopted are given below:

### 3.1 Selection of Variable

In rice production, the most important variables like the different districts, the area of cultivation in each district, the production of rice in each district for three different seasons like Samba, Kuruvai and Kodai were selected. Each season contain 2 variable area and cultivation. There are 3 seasons. Hence the total variables selected were 2 variables viz: area and production multiplied by 3 seasons gives 6 variables.

### 3.2 Data Collection

The published data by the Government of Tamil Nadu [19] was used. Training data was collected from the area and rice production of 31 districts for the year 2009-10. The test data was collected from the 5 years average of the area and rice production of 31 districts for the years 2005-06 to 2009-10.

### 3.3 Data Preprocessing

The training and test data collected had some missing data marked with 0 values. If the study uses these 0 values in computation, this 0 may divide any real number leading to infinity condition. Computer cannot compute such infinity conditions. There are many data cleaning techniques available for data preprocessing. The present study adopted the use of a global constant 0.01 in place of 0 in training and test data. This replacement avoids the computational problem of avoiding infinity during computations.

### 3.4 Architecture of ANN

The multilayer ANN architecture designed and developed in this research consists of three layers. They are:
- Input layer
- Hidden layer and
- Output layer

The architecture of ANN designed for the prediction of rice production in Tamilnadu is given in Figure 5.

#### 3.4.1 Input layer

The input layer takes the independent data pertaining to the area of cultivation and the rice production for the three seasons. These independent data are converted into sigmoid data between 0 to 1 using an activation function and the corresponding sigmoid data are represented as $x1(i)$, $x2(i)$…….$x6(i)$. The general format of the activation function is given below:

$$s(x) = \frac{1}{1+e^{-x}}$$

Where

s(x) is the sigmoid value. It varies from 0 to 1.
x is the independent input values like Area in Hectare and Rice Production in tonnes.

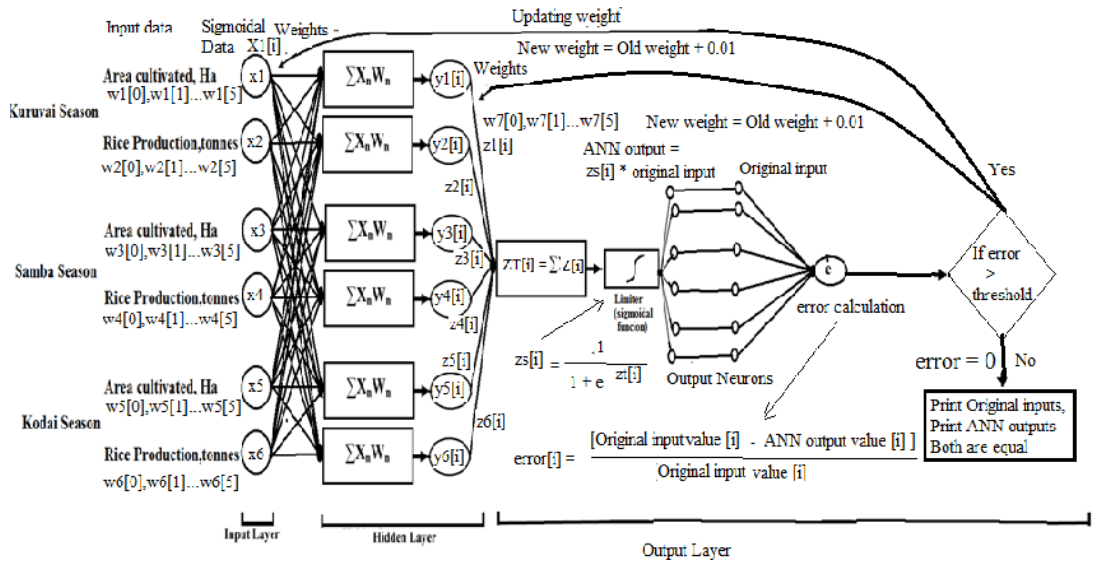

Figure 5. Architecture of ANN designed for predicting the rice production in Tamilnadu

### 3.4.2 Hidden Layer

The neurons in the hidden layers process the summation of the information received from the connections of the input layer. Then it processes the summations with its activation function and distributes the result to the next layer. This process continues down through the layers to the output layer. In the hidden layer, the sigmoid data represented as x1(i), x2(i).......x6(i) are multiplied with initially assumed weights represented as $w_i$. The initially assumed weights were from 0 to 1. Each node is multiplied with weights as follows:

$$Y_i = \sum X_i W_i + \text{Bias } (+1)$$

Where

$Y_i$ is the summation of each node $X_i$ with corresponding initially assumed weights $W_i$ plus bias. Bias used is +1. Bias is added to the summation to make the summation a number other than 0. It is essential to avoid 0 so that subsequent computations may not face division by zero (infinity). It is well discussed in data preprocessing.

### 3.4.3 Output Layer

The layer where the processed output information are computed and presented is called the output layer. Different computations performed in the output layer are given below:

| Steps | Computations performed to evaluate the performance of ANN |
|---|---|
| 1 | $Z_i$ is processed by multiplication of output $Y_i$ from the hidden layer with initially assumed weight $W_k$. There are totally six values Z(1), Z(2)……….Z(6).<br>$Z_i = Y_i * W_k$ |
| 2 | Calculation of total of all six Z values.   $ZT(i) = \sum Z_i$ |
| 3 | Convert the ZT(i) into ZS(i), that is sigmoid value from 0 to 1 using the same activation function as given below:<br>$$ZS(i) = \frac{1}{1 + e^{-ZT(i)}}$$ |
| 4 | Compute the ANN output for the six independent data like area and rice production using the formula given below:<br>ANN Output (i) = ZS(i) * Original Data (i) |
| 5 | Compute the Error between actual input data and the ANN output data using the formula given below:<br>$$Error(i) = \frac{[Original\ Data(i) - ANN\ output(i)]}{Original\ Data(i)}$$ |
| 6 | Compare the error with threshold value of $10^{-9}$. If the error is greater than threshold value then calculate the updated weights using the formula given below:<br>New updated weight = Old weight + Increment of 0.01 |
| 7 | Back propagation technique was adopted by computing the summation in the hidden layer using updated weights. It computes six sets of Yi and six sets of Zi. The total of six sets Zi is called one set of ZTi.  The conversion of  ZTi using activation function gave ZSi. The multiplication of ZSi and actual value of variables gave the ANN output. Repeat the process of feed forward and back propagation techniques until error is 0. When error is 0, print the original input and the corresponding ANN output. |
| 8 | If original independent input data and the corresponding ANN output are same, the research has predicted correctly without any deviations. This happens only when error is 0. |
| 9 | If there is difference between the original input data and the corresponding ANN output, there are some deviations. The errors were worked out to understand how far the ANN output was away from the original data. Then update the weights and follow the back propagation technique of computing summations. |

### 3.5 Supervised Learning in ANN

Learning in ANN is nothing but updating network architecture and connection weights so that the network can efficiently predict the output. Performance is improved over time by iteratively updating the weights in the network. In supervised learning, the network is provided with a correct answer (output) for every input pattern. Weights are updated to allow the network to produce answers as close as possible to the known correct answers.

$$\text{Error}(i) = \frac{[\text{Original Data}(i) - \text{ANN output}(i)]}{\text{Original Data}(i)}$$

Compare the error with threshold value of $10^{-9}$. If the error was greater than threshold value then calculate the updated weights using the formula given below:
New updated weight = Old weight + Increment of 0.01.

After finding the new weights, adopt back propagation technique to compute all summations from the beginnings until error equal to zero. It will predict exact ANN output.

## 4. RESULTS AND DISCUSSIONS

The input sigmoid values sending into the ANN and the output sigmoid values coming out from the output layers are very important aspects to predict the correct answer. Hence, they are discussed below:

### 4.1 Input Sigmoid values of independent data

The preprocessed training data was converted into sigmoid data using the sigmoid activation function and the input sigmoid values for 3 seasonal area and productions are given below:

Table 1: Frequency distribution of Sigmoid values of training data containing area of rice cultivation and its production in 2009-10

| S.No | Sigmoid value (0 to 1) | Kuruvai Season | | Samba Season | | Kodai Season | |
|------|------------------------|----------------|------------|--------------|------------|--------------|------------|
|      |                        | Area           | Production | Area         | Production | Area         | Production |
| 1    | Sigmoid value 1        | 27             | 28         | 30           | 30         | 28           | 28         |
| 2    | Sigmoid value 0.5025   | 4              | 3          | 1            | 1          | 3            | 3          |
|      | Total                  | 31             | 31         | 31           | 31         | 31           | 31         |

Table 1 show that the sigmoid value of 1 and 0.5025 were present after transformation of independent data like area and rice production using sigmoid activation function. It was found that there is a range of 1 to 4 for 0.5025 and the majority values are 1s. The total number of 0.5025 is 15 and the remaining 171 items are 1s.

### 4.2 Effect of initial weights and increments in supervised training

The initial weights assumed were within the range of 0 to 1. The weights are updated for 17 times to reach the $18^{th}$ iteration where the mean error between the original data and the ANN output is 0. Supervised training was carried out by updating the weights in every iteration so as to reduce the initial error in $1^{st}$ iteration to 0 errors in the final iteration.

The mean error between original training data and the ANN output created by the software from 1 to 18 iterations were recorded in a excel file and the graph is drawn by keeping the iteration in x axis and the mean error of Kurvai area, Kuruvai production, Samba area, Samba production and the Kodai area and Kodai productions in y axis. The graph is shown in figure 6.

Figure 6 shows the fact that the initial error computed for the ANN output from the original input data for all the variable like Kuruvai area and production, Samba area and production and the Kodai area and its production starts at 0.0000548 at the first iteration and the error of all the variables become 0 at $18^{th}$ iteration. The error reduction pattern seen from the figure 6 follows the same curve linear path. This shows that the non linearity and complexity of input data was reduced and smoothened after the transformation by the sigmoidal activation function multiplied by initial assumed weights and subsequent updated weights.

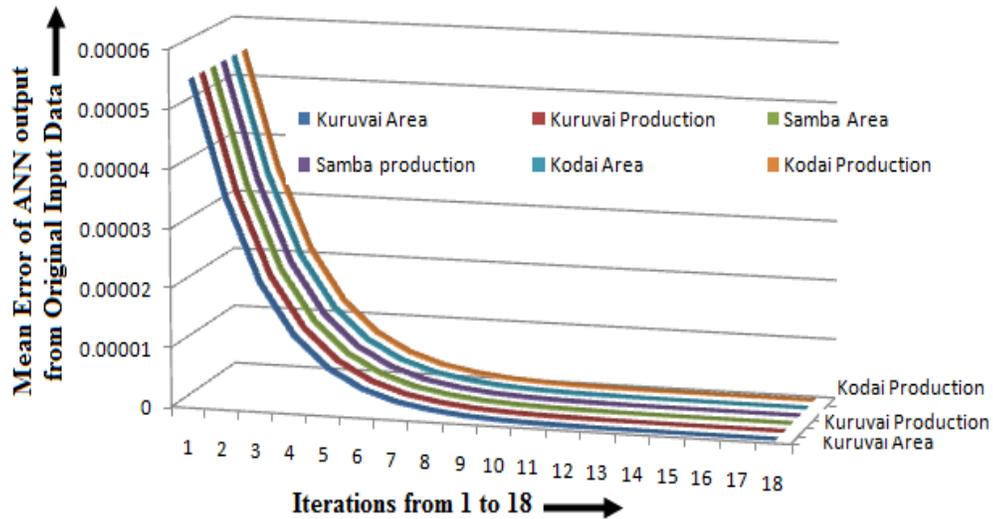

Figure 6: Mean Error between original training data and the ANN output from 1 to 18 iterations

The work of Heermann and Khazenie [15] who compared neural networks with more classical statistical methods and found that the error is distributed throughout the network which uses back propagation methods. The conclusion of Heermann and Khazenie [15] is in agreement in the present study because the error is reduced in a smooth curve linear path for all the independent variable from 0.0000548 to 0.

### 4.3 Effect of output sigmoid value for the training data from 1 to 18 iterations

It is found that the program stops its execution at $18^{th}$ iteration. The output sigmoid value for the training data starts in the order of 0.9999 for all the 31 districts in its first iterations. All districts attain the output sigmoid value of 1 at the last $18^{th}$ iteration. The output sigmoid values for all the 31 districts for all the 18 iterations were recorded in a excel file and the graph was drawn by keeping the districts in x axis and the output sigmoid value in y axis. The graph is shown in figure 7 below:

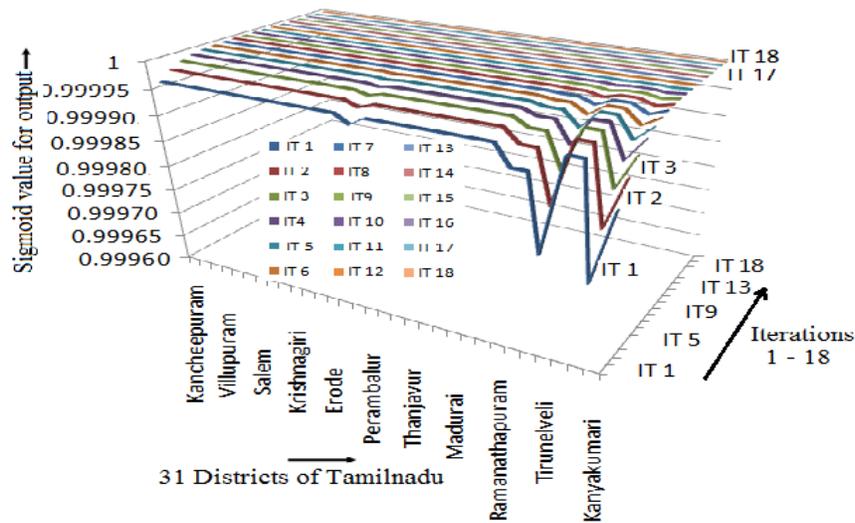

Figure 7: Effect of output sigmoid value for the training data from 1 to 18 iterations

Figure 7 shows the fact that for $1^{st}$ iteration, Karur district show a lower sigmoid value followed by Ramanathapuram and Sivagangai districts. Sivagangai has the lowest sigmoid value in the $1^{st}$ iteration. As the iterations increased from 1to 18, the sigmoid values of variable moved to 1 because the error was reduced to 0. If the sigmoid value becomes 1, which is the highest value, the graph becomes a straight line. The reason is non linear and complex data are converted into linear which is the specialty of ANN.

It is interesting to note the fact that at $13^{th}$ iteration, all the 25 districts other than districts Karur, Ramanathapuram, Virudhunagar, Sivagangai, Nilgiris and Kanyakumari got the highest output sigmoid value of 1 meaning the error was 0. To make the output sigmoid value of six districts Karur, Ramanathapuram, Virudhunagar, Sivagangai, Nilgiris and Kanyakumari to become 1 and to make the error 0, the iterations were continued further by updating the weights with increment of 0.01. It means conversion of nonlinearity to linearity continued from $13^{th}$ iterations for all the districts. The following observations were noted:

Table 2: Number of iterations and the output sigmoid value of 1 with 0 errors

| S.No | Name of Districts | No. of iterations | Output sigmoid value | Error of ANN from actual data |
| --- | --- | --- | --- | --- |
| 1 | All 25 districts excepting Karur, Ramanathapuram, Virudhunagar, Sivagangai, Nilgiris and Kanyakumari districts | $12^{th}$ | 1 | 0 |
| 2 | Karur district | $13^{th}$ | 1 | 0 |
| 3 | Ramanathapuram and Virudhunagar, | $14^{th}$ | 1 | 0 |
| 4 | Kanyakumari district | $16^{th}$ | 1 | 0 |
| 5 | Sivagangai and Nilgiris | $18^{th}$ | 1 | 0 |

(Note: output sigmoid value = 1 means error = 0)

The above facts can be well understood by seeing the straight line graph at $18^{th}$ iterations with the output sigmoid value 1 for all the 31 districts. The discussion can be summed up with the fact that the updated weight causes the summation to yield the maximum sigmoid value of +1. It means the complex and non linearity of the input data is made linear.

## 4.4 Comparison of preprocessed training and testing data with their corresponding ANN output

Table 3 shows that the original set of data is exactly matching with the ANN output. Since there are 0 errors after $18^{th}$ iterations, the ANN output is same as original input data given into the software developed and trained.

It was noted that at $12^{th}$ iteration, all the 25 districts other than districts of Karur, Ramanathapuram, Virudhunagar, Sivagangai, Nilgiris and Kanyakumari got the ANN output same as original input data. Karur district attained the got the ANN output same as original input data at $13^{th}$ iteration. Ramanathapuram and Virudhunagar districts attained the ANN output same as original input data at $14^{th}$ iteration. Kanyakumari district got the ANN output same as original input data at $16^{th}$ iteration. Finally Sivagangai and Nilgiris districts got the ANN output same as original input data at $18^{th}$ iteration.

Attaining the ANN output same as original input data means the prediction value will be exactly same as desired value. The training of ANN software is perfectly done. The software understood the complexities, non linearity and structure of training data.

Table 4 shows that the original set of test data is exactly matching with the ANN output. Since there are 0 errors after $18^{th}$ iterations during the training phase, the ANN output for test data is same as original input test data given into the software developed. The testing of ANN software is perfectly done. The software understood the complexities and structure of training data and hence it predict very correctly for the testing phase.

Similar to the ANN based rainfall prediction in Chennai [5] using back propagation neural network, which performed well both in training and testing periods, the rice prediction model developed in the study performed very well with 100% accuracy. Also, the ANN based rainfall-runoff prediction model [8] worked efficiently to predict the river runoff, the rice prediction model developed in the study performed very well with 100% accuracy.

There is 100% accuracy in prediction both in training and testing phase of the software. The software developed in the present research can be used for Tamilnadu Government's rice prediction studies.

Table 3: Comparison of preprocessed training data and the ANN output

| S.No | District | Original preprocessed testing data - - Area and production for 3 seasons | | | | | | ANN Output for testing data - Area and production for 3 seasons | | | | | |
|---|---|---|---|---|---|---|---|---|---|---|---|---|---|
| | | Kuruvai Season | | Samba Season | | Kodai Season | | Kuruvai Season | | Samba Season | | Kodai Season | |
| | | Area, Ha | Production, tonnes | Area, Ha | Production, tonnes | Area, Ha | Production, tonnes | Area, Ha | Production, tonnes | Area, Ha | Production, tonnes | Area, Ha | Production, tonnes |
| 1 | Kancheepuram | 19472.00 | 70304.00 | 63119.00 | 211983.00 | 15624.00 | 56670.00 | 19472.00 | 70304.00 | 63119.00 | 211983.00 | 15624.00 | 56670.00 |
| 2 | Thiruvallur | 39984.00 | 142745.00 | 32219.00 | 100240.00 | 13509.00 | 46658.00 | 39984.00 | 142745.00 | 32219.00 | 100240.00 | 13509.00 | 46658.00 |
| 3 | Cuddalore | 19842.00 | 66531.00 | 83546.00 | 215853.00 | 4985.00 | 16154.00 | 19842.00 | 66531.00 | 83546.00 | 215853.00 | 4985.00 | 16154.00 |
| 4 | Villupuram | 28742.00 | 92089.00 | 110654.00 | 353250.00 | 11347.00 | 36130.00 | 28742.00 | 92089.00 | 110654.00 | 353250.00 | 11347.00 | 36130.00 |
| 5 | Vellore | 9557.00 | 31802.00 | 17751.00 | 56302.00 | 18964.00 | 62439.00 | 9557.00 | 31802.00 | 17751.00 | 56302.00 | 18964.00 | 62439.00 |
| 6 | Thiruvannamalai | 22479.00 | 69618.00 | 59101.00 | 181673.00 | 33103.00 | 106394.00 | 22479.00 | 69618.00 | 59101.00 | 181673.00 | 33103.00 | 106394.00 |
| 7 | Salem | 8043.00 | 29333.00 | 17829.00 | 69336.00 | 2294.00 | 8538.00 | 8043.00 | 29333.00 | 17829.00 | 69335.00 | 2294.00 | 8538.00 |
| 8 | Namakkal | 6001.00 | 23358.00 | 7753.00 | 32294.00 | 418.00 | 1362.00 | 6001.00 | 23358.00 | 7753.00 | 32294.00 | 418.00 | 1362.00 |
| 9 | Dharmapuri | 7413.00 | 26028.00 | 12096.00 | 43806.00 | 2403.00 | 8018.00 | 7413.00 | 26028.00 | 12096.00 | 43805.00 | 2403.00 | 8018.00 |
| 10 | Krishnagiri | 7015.00 | 23332.00 | 9019.00 | 27748.00 | 1158.00 | 3677.00 | 7015.00 | 23332.00 | 9019.00 | 27748.00 | 1158.00 | 3677.00 |
| 11 | Coimbatore | 1420.00 | 5136.00 | 3071.00 | 12808.00 | 608.00 | 2946.00 | 1420.00 | 5136.00 | 3071.00 | 12808.00 | 608.00 | 2946.00 |
| 12 | Thiruppur | 258.00 | 1657.00 | 9655.00 | 41838.00 | 1597.00 | 6684.00 | 258.00 | 1657.00 | 9655.00 | 41838.00 | 1597.00 | 6684.00 |
| 13 | Erode | 10350.00 | 47157.00 | 27483.00 | 119561.00 | 1490.00 | 5435.00 | 10350.00 | 47157.00 | 27483.00 | 119561.00 | 1490.00 | 5435.00 |
| 14 | Tiruchirapalli | 6390.00 | 24114.00 | 59999.00 | 218756.00 | 3493.00 | 12164.00 | 6390.00 | 24114.00 | 59999.00 | 218756.00 | 3493.00 | 12164.00 |
| 15 | Karur | 0.01 | 42.00 | 14589.00 | 50677.00 | 452.00 | 1519.00 | 0.01 | 42.00 | 14589.00 | 50677.00 | 452.00 | 1519.00 |
| 16 | Perambalur | 1842.00 | 5783.00 | 26827.00 | 90133.00 | 1430.00 | 3712.00 | 1842.00 | 5783.00 | 26827.00 | 90133.00 | 1430.00 | 3712.00 |
| 17 | Ariyalur | 1397.00 | 4546.00 | 23536.00 | 74708.00 | 344.00 | 1208.00 | 1397.00 | 4546.00 | 23536.00 | 74708.00 | 344.00 | 1208.00 |
| 18 | Pudukottai | 905.00 | 2945.00 | 92750.00 | 204940.00 | 285.00 | 980.00 | 905.00 | 2945.00 | 92750.00 | 204940.00 | 285.00 | 980.00 |
| 19 | Thanjavur | 27368.00 | 90224.00 | 130946.00 | 350520.00 | 3567.00 | 11573.00 | 27368.00 | 90224.00 | 130946.00 | 350520.00 | 3567.00 | 11573.00 |
| 20 | Thiruvarur | 19173.00 | 62994.00 | 142823.00 | 289689.00 | 3787.00 | 13491.00 | 19173.00 | 62994.00 | 142823.00 | 289689.00 | 3787.00 | 13491.00 |
| 21 | Nagapattinam | 28363.00 | 90062.00 | 132213.00 | 250236.00 | 734.00 | 2449.00 | 28363.00 | 90062.00 | 132213.00 | 250236.00 | 734.00 | 2449.00 |
| 22 | Madurai | 8075.00 | 31990.00 | 51870.00 | 193138.00 | 3611.00 | 12419.00 | 8075.00 | 31990.00 | 51870.00 | 193138.00 | 3611.00 | 12419.00 |
| 23 | Theni | 5320.00 | 26778.00 | 9452.00 | 41540.00 | 420.00 | 1514.00 | 5320.00 | 26778.00 | 9452.00 | 41540.00 | 420.00 | 1514.00 |
| 24 | Dindigul | 2198.00 | 8169.00 | 14142.00 | 54930.00 | 2840.00 | 12081.00 | 2198.00 | 8169.00 | 14142.00 | 54930.00 | 2840.00 | 12081.00 |
| 25 | Ramanathapuram | 299.00 | 478.00 | 127446.00 | 182647.00 | 1685.00 | 5956.00 | 299.00 | 478.00 | 127446.00 | 182647.00 | 1685.00 | 5956.00 |
| 26 | Virudhunagar | 88.00 | 345.00 | 27670.00 | 95399.00 | 2445.00 | 8212.00 | 88.00 | 346.00 | 27670.00 | 95399.00 | 2445.00 | 8212.00 |
| 27 | Sivagangai | 69.00 | 235.00 | 81172.00 | 130036.00 | 47.00 | 158.00 | 69.00 | 235.00 | 81172.00 | 130036.00 | 47.00 | 158.00 |
| 28 | Tirunelveli | 22578.00 | 89786.00 | 59325.00 | 240878.00 | 5959.00 | 19011.00 | 22578.00 | 89786.00 | 59325.00 | 240878.00 | 5959.00 | 19011.00 |
| 29 | Thoothukudi | 7421.00 | 33286.00 | 10599.00 | 47633.00 | 2243.00 | 8159.00 | 7421.00 | 33286.00 | 10599.00 | 47633.00 | 2243.00 | 8159.00 |
| 30 | The-Nilgiris | 1015.00 | 3565.00 | 0.01 | 0.01 | 0.01 | 0.01 | 1015.00 | 3565.00 | 0.01 | 0.01 | 0.01 | 0.01 |
| 31 | Kanyakumari | 9758.00 | 42965.00 | 10034.00 | 42387.00 | 0.01 | 0.01 | 9758.00 | 42965.00 | 10034.00 | 42387.00 | 0.01 | 0.01 |

## 5. SUMMARY AND CONCLUSIONS

Prediction of annual rice production in all the 31 districts of Tamilnadu is an important decision for the Government of Tamilnadu so as to plan for the food security. One of the pillars of success of the Government depends on planning the availability of rice for the state. Hence, design and development of ANN system to predict rice production was successfully carried out. The system uses the popular sigmoid activation function for both converting independent data like area and production of rice and transferring the output sigmoid value back to area and production of rice. The training and testing data were collected from published report of the Government of Tamilnadu [19]. The following conclusions are drawn from the research.

1. The sigmoid value of 1 and 0.5025 were present after transformation of independent data like area and rice production using sigmoid activation function. It was found that there were few values of 0.5025 and the majority values were 1s. The total number of 0.5025 was 15 and the remaining 171 items were 1s.

2. The initial error computed for the ANN output from the original input data for the entire variable like Kuruvai area and its production, Samba area and its production and the Kodai area and its production started at 0.0000548 at the first iteration and the error of all the variables became 0 at $18^{th}$ iteration. The error reduction pattern for the entire six variables followed the same curve linear path. This showed that the non linearity and complexity of input data were reduced and smoothened after the transformation by the sigmoid activation function multiplied by weights and subsequent updated weights.

3. It was noted that at $12^{th}$ iteration, all the 25 districts other than districts of Karur, Ramanathapuram, Virudhunagar, Sivagangai, Nilgiris and Kanyakumari got the highest output sigmoid value of 1 meaning the error was 0. To make the output sigmoid value of six districts Karur, Ramanathapuram, Virudhunagar, Sivagangai, Nilgiris and Kanyakumari to become with 0 error, the iterations were continued further. Karur district attained the output sigmoid value 1 and error equal to 0 at $13^{th}$ iteration. Ramanathapuram and Virudhunagar districts attained the output sigmoid value 1 and error equal to 0 at $14^{th}$ iteration. Kanyakumari district attained the output sigmoid value 1 and error equal to 0 at $16^{th}$ iteration. Finally Sivagangai and Nilgiris districts attained the sigmoid value of 1 at $18^{th}$ iteration. Attaining sigmoid value 1 means the prediction value will be exactly same as desired value.

4. ANN architecture and software system were very well designed, developed, trained perfectly with 100% accuracy and tested perfectly with 100% accuracy in predicting the rice production for Tamilnadu. Hence, the research was completed with 100% accuracy in prediction.

## ACKNOWLEDGEMENTS

The PhD scholar first and foremost thank the almighty God for his abundant blessings and protection during my studies and research work. I am highly indebted to Dr. K. Baskaran, my research Director and Associate Professor, Dept. of Computer Science & Engineering and Information Technology, Government College of Technology, Coimbatore for his involvement, encouragement, suggestions and constructive criticism to make the research a grand success. I also thank the members of my research committee Dr. Ebenezer Jeyakumar, Director (Academic), Sri Ramakrishna College of Engineering, Vattamalaipalayam, Coimbatore and Dr. M. Hemalatha, Head of the Department of Software systems, Karpagam University Coimbatore for their suggestions and guidance to make my work more applied to the society.

## AUTHORS: SHORT BIOGRAPHY

1) Mr. S.Arun Balaji  B.E  and M.Tech (Information Technology)
PhD Scholar Computer Science & Engineering, Karpagam University
Coimbatore, India.
Teaching Experience: Total 4 years (2 years in India, 2 years abroad)
Email: arunbalaji1983@yahoo.co.in

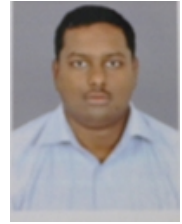

2) Dr. K. Baskaran PhD; Associate Professor (CSE & IT)
Government College of Technology, Coimbatore– 641013, India
B.E, M.E and PhD: Computer Science & Engineering
Email:   baski_101@yahoo.com
Mobile: +91-9443661901

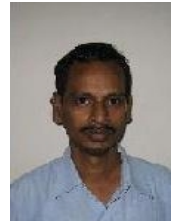